\pdfoutput=1

\documentclass[11pt]{article}

\usepackage{EMNLP2022}

\usepackage{times}
\usepackage{latexsym}

\usepackage[T1]{fontenc}

\usepackage[utf8]{inputenc}

\usepackage{microtype}

\usepackage{inconsolata}

\usepackage{amsmath}
\usepackage{amssymb}
\usepackage{amsthm}
\usepackage{graphicx}
\usepackage{mathtools}
\usepackage{listings}
\usepackage{tikz}
\usepackage{siunitx}
\usepackage{tipa}
\usepackage{float}
\usepackage{bbm}
\usepackage{breqn}

\usepackage{caption}
\usepackage{subcaption}

 \usepackage{todonotes}

\makeatletter
\newcommand*\iftodonotes{\if@todonotes@disabled\expandafter\@secondoftwo\else\expandafter\@firstoftwo\fi}  %
\makeatother

\DeclareMathOperator{\simm}{S}
\DeclareMathOperator{\meann}{mean}

\usepackage{booktabs, multirow} %

\newcommand{\wlab}[1] {\texttt{#1}}

\title{

Why is Winoground Hard?\\
Investigating Failures in Visuolinguistic Compositionality}

\author{Anuj Diwan$^{*,\diamondsuit}$, \ Layne Berry$^{*,\diamondsuit}$, \ Eunsol Choi$^{\diamondsuit}$, \ David Harwath$^{\diamondsuit}$, \ Kyle Mahowald$^{\heartsuit}$ \\
  	$^\diamondsuit$Department of Computer Science \ \  $^\heartsuit$Department of Linguistics \\ The University of Texas at Austin \\
  \texttt{\{anuj.diwan, layne.berry, eunsol, harwath, mahowald\}@utexas.edu}}

\begin{document}
\maketitle
\begin{abstract}

Recent visuolinguistic pre-trained models show promising progress on various end tasks such as image retrieval and video captioning. Yet, they fail miserably on the recently proposed Winoground dataset~\citep{winoground}, which challenges models to match paired images and English captions, with items constructed to overlap lexically but differ in meaning (e.g., ``there is a mug in some grass'' vs. ``there is some grass in a mug''). By annotating the dataset using new fine-grained tags, we show that solving the Winoground task requires not just compositional language understanding, but a host of other abilities like  commonsense reasoning or locating small, out-of-focus objects in low-resolution images. 
In this paper, we identify the dataset's main challenges through a suite of experiments on related tasks (probing task, image retrieval task), data augmentation, and manual inspection of the dataset. 
Our analysis suggests that a main challenge in visuolinguistic models may lie in fusing visual and textual representations, rather than in compositional language understanding. We release our annotation and code at \url{https://github.com/ajd12342/why-winoground-hard}.

\end{abstract}
{\let\thefootnote\relax \footnotetext{*Co-first authors contributed equally. Order determined by coin flip.} }
\section{Introduction}

\begin{figure}
    \centering
    \includegraphics[width=1\columnwidth]{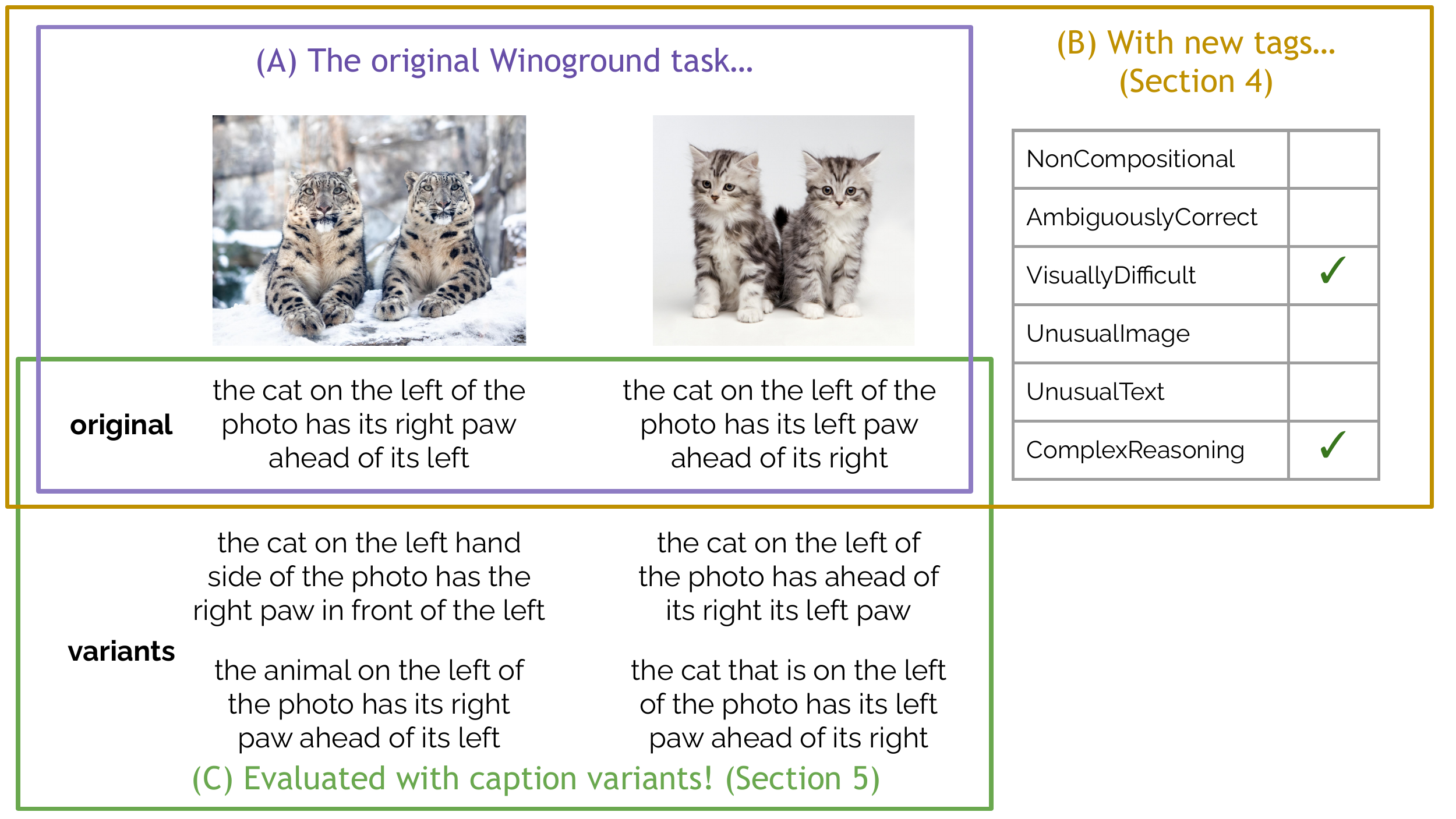}
    \caption{Extending the (A) original Winoground task, which presents a minimal semantic pair of image captions and their corresponding images, we (B) create new fine-grained tags, identify their main challenges, and evaluate performance separately on each subcategory (Section~\ref{sec:taxonomy}); further, we (C) also create textual variants of the original captions where they are no longer minimal semantic pairs. Models are still unable to succeed on the Winoground task (Section~\ref{sec:augmentations}) when given such linearly separable pairs.}
    \label{fig:wingroundex}
\end{figure}

Despite the success of large pretrained transformer models on a wide variety of tasks, the extent to which they are \textbf{compositional}  \citep[e.g.,][]{kim-linzen-2020-cogs,soulos-etal-2020-discovering,hewitt-manning-2019-structural,sinha-etal-2021-masked,clouatre-etal-2021-mlmlm} and  \textbf{grounded} \citep{bender-koller-2020-climbing,bisk-etal-2020-experience} is debated. Taking compositionality and groundedness as key desiderata, the recent Winoground dataset~\citep{winoground} provides a clever way to test multimodal vision and language models. 
Given two images and two captions, the goal is to pair them correctly. 
The key insight is that, inspired by the Winograd schema~\citep{10.5555/3031843.3031909}, the two captions contain the same set of words/morphemes, only in a different order.
Figure~\ref{fig:wingroundex}(A) shows a representative example.

Pretrained multimodal transformer models \citep[e.g.][]{clip,lxmert,uniter} have achieved impressive performance in multimodal tasks like image retrieval, image captioning, and visual question answering, as measured on a variety of datasets \citep[e.g.,][]{johnson2017clevr,suhr-etal-2017-corpus,bitton-etal-2021-automatic}.
But, on Winoground, they all fall down: not one performs meaningfully better than random chance---despite the fact that humans can easily do the task.

Citing evidence from \citet{sinha-etal-2021-masked} that large language models don't need word order information to do well on tasks \citep[see also][]{sinha-etal-2021-unnatural,hessel-schofield-2021-effective,pham2021out,gupta2021bert,oconnor-andreas-2021-context}, the Winoground authors suggest that models track word co-occurrences, thus giving ``the
illusion of an understanding of word order'' without actually achieving that understanding \citep{winoground}.
Indeed, given that information about semantic meaning can be uncovered without word order information \citep[e.g.,][]{papadimitriou-etal-2022-classifying-grammatical} and that seemingly syntactic and semantic tasks can be solved with lexical heuristics \citep[e.g.,][]{mccoy-etal-2019-right,sinha-etal-2021-unnatural}, Winoground failures may offer another evidence that language models solve complex tasks in a relatively superficial way.

To assess this possibility, we examine the Winoground task and conduct a series of novel experiments on the dataset, testing three models (CLIP, \citealt{clip}; UNITER, \citealt{uniter}; LXMERT, \citealt{lxmert}) that reflect three broad categories of Transformer-based vision-and-language architectures.
First, we test these models on the more general (and standard) text-to-image and image-to-text Recall@K task using the Winoground images and captions. %
Some models fail at this simpler task, suggesting that failure on Winoground may not be just because of a failure in semantic composition but due to broader difficulty with atypical images in the dataset.
We show
that even fine-tuning probes specifically on Winoground  does not help, implying a potential absence of information necessary to succeed at the task.

Second, to understand what the source of failure might be, we develop a new taxonomy of Winoground examples consisting of six classes (Section~\ref{sec:taxonomy}). 
Our taxonomy reflects various abilities required to solve the task, and model's performances vary significantly among our classes.
Given an explosion of interest in testing image generation models (e.g., DALL-E 2 and Imagen) on their compositional ability \citep[e.g.,][]{marcus2022very}, Winoground can be a crucial benchmark, which motivates the need for a deeper analysis of its properties; prior work has performed such deep analyses for other benchmarks~\citep{alt-etal-2020-probing,https://doi.org/10.48550/arxiv.2208.11695,luo-etal-2022-find}. 
We tag every example in Winoground with our scheme (see Appendix~\ref{sec:appendixnewtags} for full reporting) and provide a performance breakdown of models. Figure~\ref{fig:wingroundex}(B) shows these tags.
We show high variability in performances based on our proposed tags and observe low performance on tags that are challenging for reasons beyond compositional language understanding (e.g., a low-res version of the image simply lacks the visual detail necessary for answering the question). %
Thus, we conclude that not all Winoground items test what they aim to, and identify a subset of 171 items which directly measure compositionality. 

Third, we run a series of probing experiments to better understand whether the failure arise because of failures in visual discrimination, in linguistic compositionality, or in the fusion of vision and language.
Specifically, we augment the original captions with a set of textual variants~\cite{dhole2021nlaugmenter}.
While these textual variants are indeed highly separable in embedding space, using them fails to improve the task performance. Figure~\ref{fig:wingroundex}(C) shows these textual variants.

Taken together, our results suggest that failures found on Winoground reflect meaningful model failures.
While some Winoground items may be ill-suited to evaluate compositionality, even the most straightforward items pose a challenge. Our evidence suggests that the source of these robust failures lies in fusing visual and linguistic information, not strictly in complex language understanding. 
We hope our analysis will help future endeavors in interpreting emerging models' Winoground performance.

\section{Background}

The Winoground dataset \citep{winoground} contains 400 items (each consisting of two image+text pairs with overlapping lexical content).
The items were categorized linguistically based on whether the text swaps an object, a relation, or both. 
The items were further categorized based on if they involved: a $\texttt{Pragmatics}$ tag indicating non-literal/pragmatic reasoning required, a $\texttt{Symbolic}$ tag indicating reasoning about something in symbolic space (e.g.,  children's drawing), and a $\texttt{Series}$ tag (indicating whether the items come from the same, as opposed to from unrelated, photos). 

Evaluated models see one image/caption pair at a time for a given item, where an item consists of two pairs: $I_0$ and its paired caption $C_0$, and  $I_1$ and \textit{its} paired caption $C_1$. They then compute an Image Score, Text Score, and Group Score (by scoring each item as either 1 or 0 and then aggregating).
For a given pair, the Image Score is 1 if and only if for image $I_0$ a higher score is assigned to caption $C_0$ than $C_1$ \textit{and} for $I_1$ a higher score is assigned to $C_1$ than $C_0$.
Similarly, the Text Score is 1 if and only if for text $C_0$ a higher score is assigned to $I_0$ than $I_1$ (\textit{and} vice versa for $C_1$). 
Thus, for both the Image and Text Score, random chance is $1/4$.
An item's Group Score is 1 if and only if both its Text and Image Scores are 1. The random chance for Group Score is $1/6$. %

\section{Relaxing Winoground Constraints}
These metrics are relatively harsh in two respects. First, they require perfect matching between the images and captions, implicitly evaluating an unusual variant of Recall: Recall @ 1 over $2$ candidates (i.e., the best image must be ranked first between the two candidates). Further, they do not allow any adaptation to the task (i.e., zero-shot transfer is required). We therefore relax each of these two constraints in turn.

\subsection{Recall at $k>1$}\label{sec:retrieve}

\paragraph{Setting}
We evaluate using a standard Recall at $k$ (R@$k$) metric for retrieval, which asks whether the correct caption (for Image-to-Text, or I2T, retrieval) or image (for Text-to-Image, or T2I, retrieval) is present in the top $k$ candidates as ranked by the model. We consider R@1, R@2, R@5, and R@10 for CLIP, UNITER, and LXMERT (see  Appendix~\ref{app:eval_traditional_methods} for further model details). 

Crucially, R@1 requires discriminating both within semantic minimal pairs and between unrelated Winoground items, while all other metrics can be solved without having to differentiate the semantic minimal pairs (i.e., for R@2, the model can simply return both relevant items).

\paragraph{Methods}
Each model is used to compute a similarity score for all $800\times800$ possible pairs of any image from Winoground with any caption from Winoground. As in the original Winoground methodology, we do not finetune the models. In I2T retrieval, we score each image in turn and retrieve the top $k$ highest-ranked captions. In T2I retrieval, we score each caption in turn and retrieve the $k$ highest-ranked images. In either case, we then compute R@$k$ as the percentage of image or caption prompts for which the correct match is among the top $k$ candidates.

\paragraph{Results} Table \ref{tab:t2i} presents the results. 
CLIP performs well on the less harsh R@5 and R@10 metrics, while LXMERT performs poorly across all values of $k$, with UNITER's performance falling about halfway in between. 
Since neither UNITER nor CLIP clearly outperforms the other on the Winoground metrics \citep{winoground}, the stark difference in overall R@$k$ that we see between them here is surprising. 
One plausible explanation for this pattern is that LXMERT sees only about $180$K unique images during pretraining (despite seeing between $9$M and $10$M captions), while UNITER sees about $4.2$M and CLIP sees $400$M. 
We hypothesize that CLIP's larger training set size means that it can more easily adapt to unusual texts and images. Our results suggests that while the strict evaluation metric of Winoground leaves the three models at similar baseline performance, they clearly exhibit different levels of understanding Winoground captions in easier setting.

\begin{table}%
\footnotesize
    \centering
    \begin{tabular}{ccccccc}
        \toprule
             & \multicolumn{2}{c}{\textbf{CLIP}} & \multicolumn{2}{c}{\textbf{UNITER}} & \multicolumn{2}{c}{\textbf{LXMERT}}\\ 
             
        & \textbf{T2I} & \textbf{I2T} & \textbf{T2I} & \textbf{I2T} & \textbf{T2I} & \textbf{I2T} \\
        \midrule
        \textbf{R@1} & 32.9 & 27.4 & 20.1 & 16.4 & 5.9 & 3.4 \\

        \textbf{R@2} & 54.4 & 47.9 & 31.4 & 28.7 & 10.1  & 6.9 \\

        \textbf{R@5} & 72.4 & 65.9  & 45.0  &  43.8 & 18.6  & 12.0 \\

        \textbf{R@10} & 81.3 & 78.4 & 55.3 &  55.4 &  26.5   & 15.6 \\
        \bottomrule
    \end{tabular}
    \caption{Text to Image (T2I) and Image to Text (I2T) Retrieval over Winoground.}
    \label{tab:t2i}
\end{table}

\subsection{Task Adaptation}\label{sec:probe1}

\citet{winoground} evaluate models on Winoground zero-shot (with no fine-tuning to allow it to adapt to the task) and in such a way that the model is fed one caption $T_i$ and one image $I_i$ at a time (meaning, in choosing the best image match for $T_0$, it does not get to simultaneously compare $I_0$ and $I_1$ in the way that a human does).
To test whether performance is helped by addressing both factors, we train \textbf{probes to select between two concatenated cross-modal embeddings as to which represents the better match for a given reference item.}
This amounts to a binary classification task, where the output is $0$ if the first embedding is a better match, or $1$ if the second is better. 

\paragraph{Methods}
We first divide the 400 Winoground items into 300 for training and 100 for testing. Stratified sampling is used to ensure that the original ratios of each Winoground tag (\texttt{Pragmatic}, \texttt{Symbolic}, etc.) are preserved in each subset. Our probes are 4-layer MLPs with a hidden dimension of 1024 trained for 200 epochs on the embeddings of the training items. We consider the \textit{Pooled Output} embeddings produced by both UNITER and LXMERT, which are generated by applying a linear projection and Tanh activation to the hidden state of the CLS token at the last layer of each model; these are the embeddings used to predict similarity scores in the retrieval setting.
Two variants of each probe are learned: one which picks between embeddings of the same caption with two different images (roughly corresponding to Text Score or I2T retrieval), and one which picks between embeddings of the same image with two different captions (roughly corresponding to Image Score or T2I retrieval). We report additional methodological details in Appendix~\ref{app:finetune}.

In addition to our target task of picking the correct match within each Winoground item, we train another set of probes which learn a control task. For our control task, we randomly pick $50\%$ of the training items and $50\%$ of the testing items and flip their labels, then train the probes the same way. All probes are trained and evaluated 11 times with different random seeds, and the min and max score across trials is recorded.%

\paragraph{Results}
Probing results are reported in Table \ref{tab:pooled_probe}. 
None of the probes achieve an appreciably higher accuracy than either chance ($50\%$) or the control on the test set (although the UNITER text and image probe test accuracies trend somewhat higher than the UNITER control accuracies). This implies that the representations produced by LXMERT or UNITER may not contain the information required to succeed on Winoground, although it is possible that a different probe design or probing technique may be able to extract such information.

\begin{table}%
\footnotesize
    \centering
    \setlength{\tabcolsep}{3pt}
    \begin{tabular}{ccccc}
        \toprule
             & \multicolumn{2}{c}{\textbf{LXMERT}} & \multicolumn{2}{c}{\textbf{UNITER}}\\ 
             \cmidrule(lr){2-3}\cmidrule(lr){4-5}
        & \textbf{Text} & \textbf{Image} & \textbf{Text} & \textbf{Image}  \\
        \cmidrule(lr){2-5}
        \textbf{Target (Test)} & 49.0-54.5 & 48.5-51.8 & 53.5-59.5 & 52.2-55.0 \\
                \textbf{Control (Test)} & 42.2-58.5 & 48.2-57.2 & 44.0-54.8 & 44.2-54.5 \\
        \bottomrule
    \end{tabular}
    \caption{Training and test accuracies over pooled outputs results for 4-layer probe. For each measure, we report min and max accuracies over $11$ runs with random seeds.}
    \label{tab:pooled_probe}
\end{table}

\section{Characterizing the Challenges Presented by Winoground Items}\label{sec:taxonomy}

The results of our more traditional evaluation suggest that the Winoground text/image pairs are, even without focusing on semantic minimal pairs, interestingly different from other visuolinguistic datasets.
In this section, we seek to characterize what makes the Winoground task challenging. 
See Appendix~\ref{app:tags} for details on our annotation method and Table~\ref{tab:tags} for tag to dataset item mappings. We introduce our taxonomy below, and present examples of each new tag in Figure~\ref{fig:biggrid}.

\subsection{Potentially Easy Pairs}

\paragraph{\wlab{NonCompositional}} While these items are textual minimal pairs, they are actually not semantically compositional variants of one another. This may be because the swapped words appear in a compound (e.g. ``banana split'' in WG \#133, ``downfall'' in WG \#325), because they are part of an idiom (e.g. ``fishing for compliments'' in WG \#333), or because they are two different lexemes exhibiting polysemy. 
Items with this tag do not require compositional reasoning to resolve, since they don't contain the same semantic entities.

\subsection{Potentially Difficult Pairs: In-Domain}

We identify two challenging categories of examples that are in-domain, but involve additional challenges beyond visual or linguistic understanding. %

\begin{figure*}
    \centering
    \includegraphics[width=\textwidth]{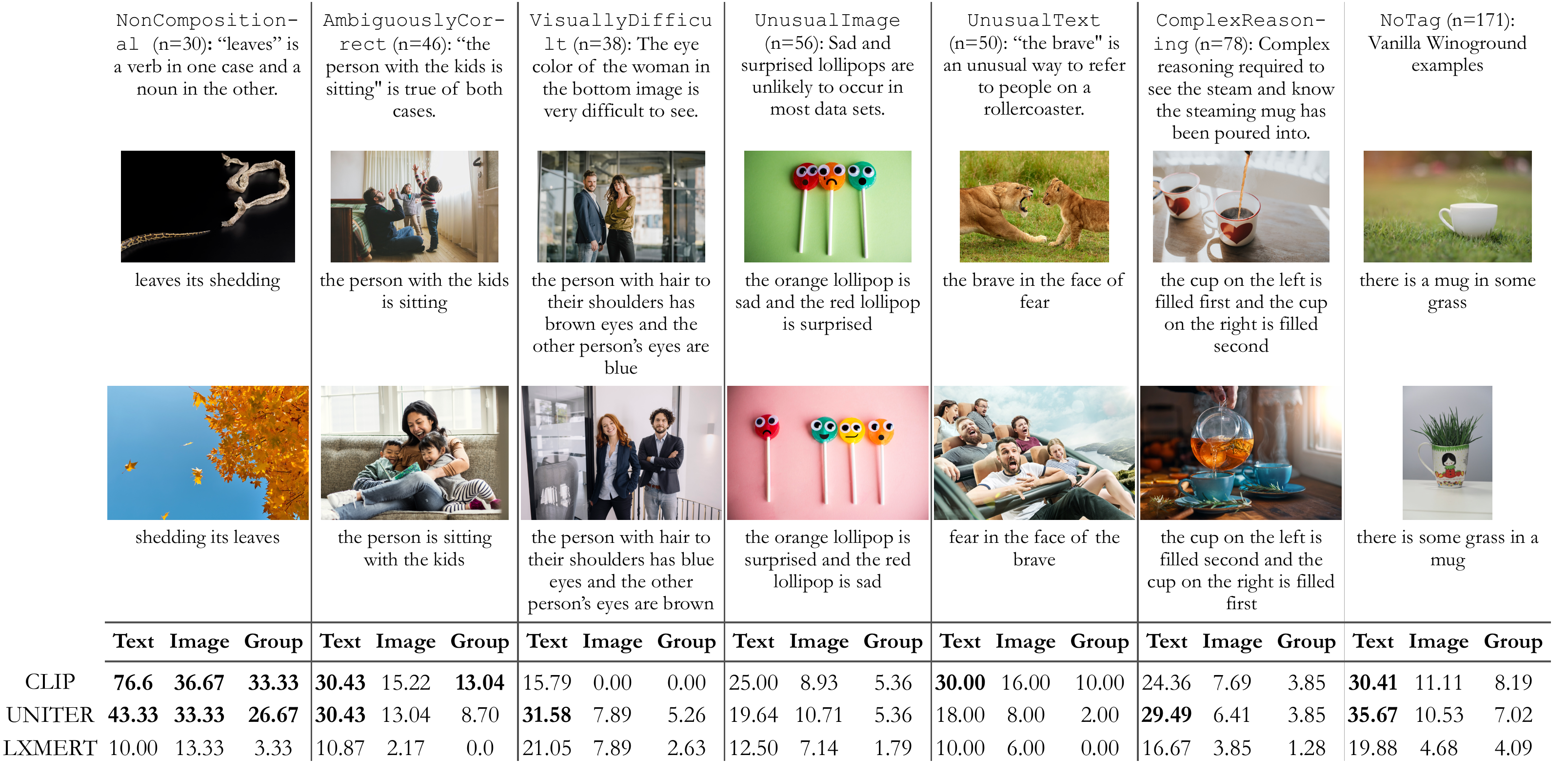}\vspace{-0.5em}
    \caption{A taxonomy of Winoground schemes, with scores on CLIP in the bottom row for Text/Image/Group score respectively and with above-chance performance in bold.}\vspace{-0.5em}
    \label{fig:biggrid}
\end{figure*}

\paragraph{\wlab{AmbiguouslyCorrect}}
These items can be resolved when both images and both captions are considered together, but when considered separately, at least one of the captions is either a correct description of both images or not quite a correct description of either.
SOTA Transformer-based VL models are trained to distinguish valid captions from invalid captions, but not to select the best caption from a set of valid candidates.
Humans, while capable of making such fine-grained judgments, were queried differently than models in \citet{winoground}: rather than rating the quality of an image-caption pair along a continuum (analogous to models' similarity scores), humans were asked for a binary judgment. 
Even a perfect respondent, if asked to evaluate some of these image/text pairs in isolation (without seeing the competitor pair), could receive zero Winoground scores since the correct answer is only discernible when both competitors are present.

\paragraph{\wlab{VisuallyDifficult}}
For items given this tag, at least one element required to correctly sort the images is small, blurry, in the background, out-of-focus, indistinct, blends with the background, or otherwise difficult to detect. 
Since most VL models have low input image resolution, %
they may simply be unable to detect visual elements which are key to resolving these Winoground items.

\subsection{Potentially Difficult Pairs: Out-of-Domain} 

We also identify three kinds of out-of-domain reasoning required to solve the Winoground task: either because the image is unusual, the text is unusual, or because they require extensive real-world knowledge or reasoning ability. 
While humans can adapt to out-of-domain tasks and it is desirable to build systems that can as well, this goes beyond mere compositionality.

\paragraph{\wlab{UnusualImage} and \wlab{UnusualText}}
Items that we tag \wlab{UnusualImage} have at least one image which is either entirely unrealistic or highly unusual and therefore likely out-of-distribution for most VL models. \wlab{UnusualText} captions may be difficult for models to resolve because they include a misspelled word (only found in WG \#327); because non-standard capitalization is used in one of the captions (found in $3$ Winoground items); because they're ungrammatical in Standard English (found in $5$ Winoground items); or, most commonly, because the wording of the caption is awkward. 
These may be descriptions a human would be highly unlikely to generate (e.g. WG \#10, which captions an image of a boat ``the water rests below the sail'') or phrases which are difficult to parse.

\paragraph{\wlab{ComplexReasoning}}
This category encompasses any item which requires common-sense reasoning or world knowledge to resolve. This may be numerical reasoning, as in WG \#396 (which requires counting to 3 and 8 and identifying even and odd numbers); understanding of non-English languages, as in WG \#298 (which requires the model to first perform OCR, then understand French text sufficiently to know ``chaud'' is hot and ``froid'' is cold); recognition of scientific terminology, as in WG \#303 (which requires the model to know that a lizard is cold-blooded while a polar bear is warm-blooded); or causal inference regarding the ongoing events depicted, as in the example in Figure~\ref{fig:biggrid}.

\subsection{Results on New Tags}
We compare Text, Image, and Group Score (as in \citealt{winoground}) over the splits corresponding to each of our new tags, as well as on the 171 items which don't receive any tag. 
Results for CLIP, LXMERT and UNITER are reported in the table in Figure \ref{fig:biggrid}, with scores beating random chance in bold.

As predicted, all of the potentially difficult tags are harder than the \wlab{NonCompositional} tag, in some cases strikingly so.
For CLIP, performance on the 38 \wlab{VisuallyDifficult} tags is actually 0 for the Image and Group Score metrics, suggesting that for at least some items there may just not be sufficient visual information available for the model to make an accurate judgement.
CLIP performs above random chance on all three metrics only for the \wlab{NonCompositional} tag, which tests the models' response to highly similar texts without testing their compositional reasoning. 
CLIP also performs better on the \wlab{AmbiguouslyCorrect} tag than it does on the full dataset: it appears that CLIP is able to discriminate between multiple valid or multiple invalid captions for an image to some extent, even if distinguishing between multiple valid or multiple invalid images for a caption remains out of reach. 
CLIP's much higher scores on the \wlab{NonCompositional} split compared to all other splits, including the \wlab{NoTag} split, implies that it is compositional reasoning in particular which makes Winoground so difficult, at least for the CLIP model evaluated here.

For LXMERT, the \wlab{AmbiguouslyCorrect} and \wlab{UnusualText} tags appear to be particularly challenging, and the \wlab{VisuallyDifficult} tag doesn't appear to present much of a problem.
However, it's worth noting that all LXMERT scores are below random chance--we therefore cannot be certain that any particular score difference is not a coincidence. LXMERT's failure to perform coarse-grained retrieval over the full Winoground dataset makes it unsuprising that it cannot correctly match even the potentially easy \wlab{NonCompositional} tag.%

UNITER is able to beat random chance on Text Score in all cases except for the  \wlab{UnusualImage} and \wlab{UnusualText}, suggesting that out-of-domain samples are a particularly salient challenge for UNITER.
In terms of Image and Group Score, UNITER is only able to beat random chance on the \wlab{NonCompositional} tag. This again implies that it is not just textual minimal pairs that cause catastrophic failure, but specifically textual \textit{and semantic} minimal pairs. 

\section{Generating Non-Minimal Winoground Data with Textual Variants}
\label{sec:augmentations}

In this section, we look in depth at whether the minimal textual pairs are simply not sufficiently distingushable with existing vision and language models. 
That is, at the level of text, does the model not understand that ``grass in the mug'' is distinguishable from ``mug in the grass''? Or is the problem instead that the images are not distinguishable---or that the fusion of the visual and linguistic information is too difficult?

To tease apart these hypotheses, we run experiments using \textit{caption variants}: we modify each caption in each Winoground item so that the captions are no longer minimally contrastive.
We obtain caption variants by using $9$ manually selected augmentation strategies from NLAugmenter~\citep{dhole2021nlaugmenter} and categorize them by the type of modification they make (see Table~\ref{tab:augmentations} for an example). For a given Winoground item $(I_0, I_1, T_0, T_1)$, the $n$ caption variants are denoted by $T_{0,0:n-1}$ and $T_{1,0:n-1}$. For more details about these augmentation strategies, refer to Appendix~\ref{app:variants}.

We first investigate the separability of textual variants of $T_0$ from textual variants of $T_1$ in model embedding space for the three models (LXMERT, UNITER, CLIP) in Section~\ref{sec:probe2}. Then, we test whether providing models access to textual variants helps performance on the Winoground task in Section~\ref{sec:probe3}. Finally, we analyze the ability of models to distinguish the right caption conditioned on its textual variant in Section~\ref{sec:probe4}.

\begin{table*}
    \footnotesize
    \centering
   \begin{tabular}{ p{4.1in} | p{2.0in}}
        \textbf{Augmentation} & \textbf{Example Sentence} \\  \midrule
     \textbf{Original Sentence (1)}: no changes from Winoground & a human viewing a cat on a screen \\ \midrule
    \textbf{Hyponyms (2)}: replace noun with hyponym, from CheckList~\citep{ribeiro-etal-2020-beyond}
 &  a human viewing a \textcolor{blue}{lion} on a screen\\ 
           \textbf{Hypernyms (2)}: replace noun with hypernym, from CheckList~\citep{ribeiro-etal-2020-beyond} & a human viewing a \textcolor{blue}{device} on a screen\\ 
                    \textbf{SynonymSubstitution (3)}: replace word with WordNet~\citep{miller1998wordnet} synonym &  a human \textcolor{blue}{view} a cat on a screen \\ 
                    \textbf{Slangificator (3)}: replaces a word with a slang word from a curated word list  & a human viewing a \textcolor{blue}{moggie} on a screen \\ 
                    \textbf{Backtranslation (1)}: translate to German and back using FSMT~\citep{ng-etal-2019-facebook} & a human \textcolor{blue}{looking at} a cat on a screen \\ 
                        \textbf{DiverseParaphrase (3)}: diverse paraphrases \citep{kumar-etal-2019-submodular} &\textcolor{blue}{what is it like to look at} a cat on screen \\ 
                    \textbf{ProtAugmentDiverseParaphrase (5)}: diverse paraphrases \citep{dopierre-etal-2021-protaugment} & a \textcolor{blue}{person who looks at} a cat on a screen \\ 
     \textbf{Syntactic (3)}: use hardcoded syntactic rules to generate text with a new word order but same semantics using the AllenNLP of SRL BERT~\citep{https://doi.org/10.48550/arxiv.1904.05255}  & a human viewing \textcolor{blue}{on a screen a cat} \\ 
                    
  \end{tabular}
    \caption{Text augmentations (modifications from the original sentence colored in blue) and examples. The parenthetical number states the maximum number of variants we produced for each augmentation type. For a given caption, if an augmentation did not apply (either because it reproduced the original sentence or produced the empty string), it was not included.}
    \label{tab:augmentations}
\end{table*}

\subsection{Separability of Caption Variants}\label{sec:probe2}

Our core question in this experiment is whether textual variants of $T_0$ and textual variants of $T_1$ are  effectively partitioned in each model's embedding space. 
If semantic differences aren't captured by the language branch, then no matter how well fine-grained semantics are extracted from images and no matter how well text semantics and image semantics are aligned, these models cannot be expected to succeed on Winoground. On the other hand, if there's a clear linear division between the caption groups, then the semantic distinctions between $T_0$ and $T_1$ are already easily retrievable from a model's text branch, and the model's overall failure cannot be resolved by improvements to its ability to discriminate text.

For each Winoground item, we construct four sets of CLS @ $l$ embeddings (the embedding for the \texttt{[CLS]} token at layer $l$): variants of caption 0 conditioned on image 0 ($E(I_0,T_{0,0:n-1})$), variants of caption 0 conditioned on image 1 ($E(I_1,T_{0,0:n-1})$), variants of caption 1 conditioned on image 0 ($E(I_0,T_{1,0:n-1})$), and variants of caption 1 conditioned on image 1 ($E(I_1,T_{1,0:n-1})$).  We fix the image input, and compare the target task of distinguishing variants of caption 0 from variants of caption 1 with a control task where variants of both captions are randomly assigned to one of two arbirary sets.

Separately, for each of the 400 Winoground items with textual variants, we use a Linear Support Vector Classifier probe to measure separability, with hyperparameter $C=100$ to prioritize complete separation over margin width. We obtain two key measures: the binary variable of whether the sets are linearly separable (true if and only if every variant is correctly labeled by the learned probe), and the width of the discovered margin (computable by $M=2/\Vert w\Vert$). We train one SVC over CLS @ $l$ embeddings for each combination of task, layer, Winoground item, model, and (for LXMERT and UNITER) which image is input alongside the text, then average across items and images to analyze high-level trends.

\begin{figure*}
    \centering
    \begin{subfigure}[b]{.55\columnwidth}
        \includegraphics[width=\columnwidth]{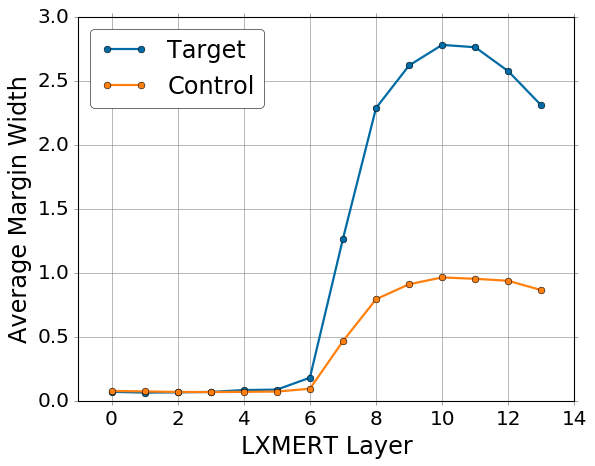}
    \end{subfigure}
    \begin{subfigure}[b]{.55\columnwidth}
        \includegraphics[width=\columnwidth]{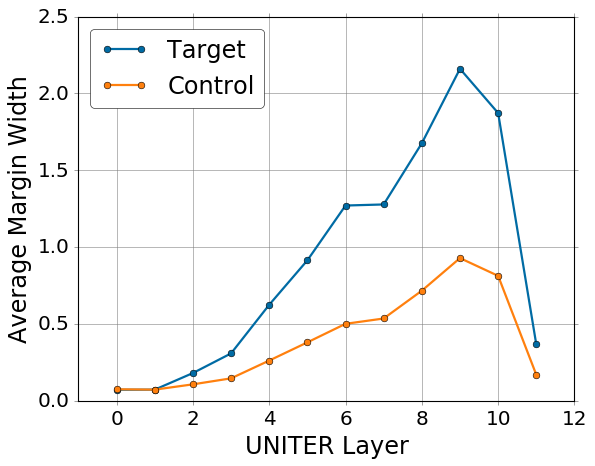}
    \end{subfigure}
    \begin{subfigure}[b]{.55\columnwidth}
        \includegraphics[width=\columnwidth]{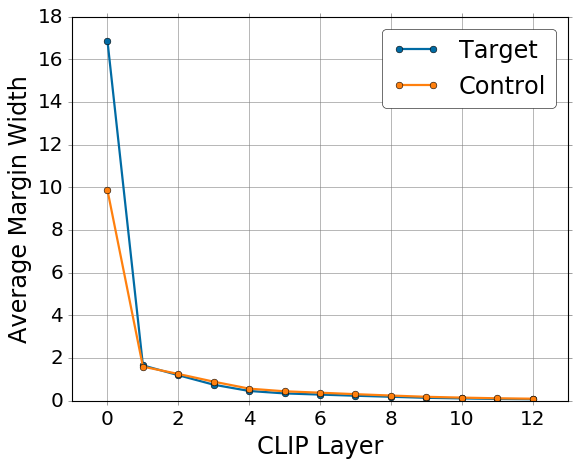}
    \end{subfigure}
    \begin{subfigure}[b]{.55\columnwidth}
        \includegraphics[width=\columnwidth]{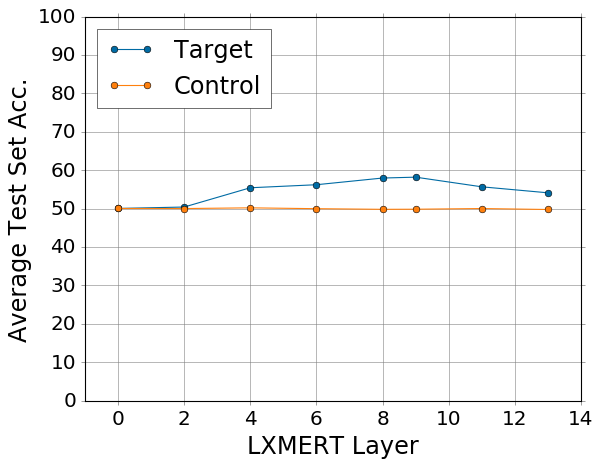}
    \end{subfigure}
    \begin{subfigure}[b]{.55\columnwidth}
        \includegraphics[width=\columnwidth]{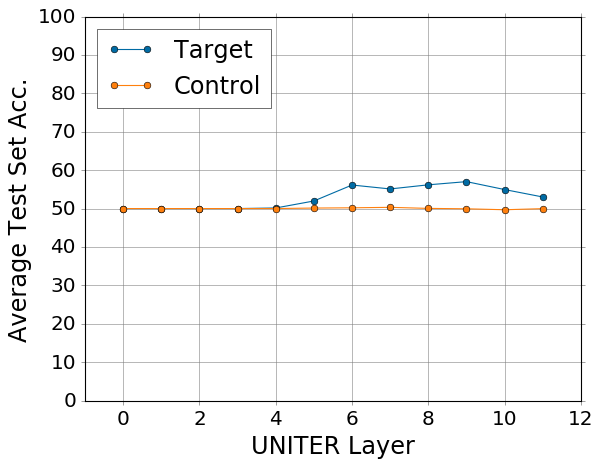}
    \end{subfigure}
    \begin{subfigure}[b]{.55\columnwidth}
        \includegraphics[width=\columnwidth]{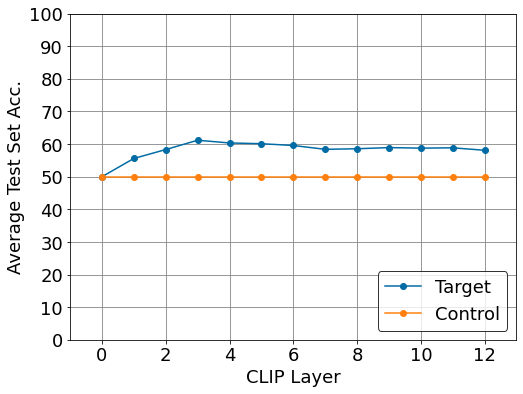}
    \end{subfigure}
    \caption{Top: Average SVC Probe Margin Width Across Model Layers for LXMERT, UNITER, and CLIP. Bottom: Test Set Accuracy of a 4-layer text-only probe over [CLS] Embeddings Across Model Layers for LXMERT, UNITER, and CLIP.}
    \label{fig:svc}
\end{figure*}

The first row of Figure~\ref{fig:svc} shows the results. For LXMERT, we find that embeddings only become separable with a margin size of at least $1.0$ at layer 6 and remain separable for the rest of the layers. The introduction of cross-modal attention at layer 9 is followed by a slowing of margin growth and an eventual decrease. Control sets are much less likely than target sets to be linearly separable. For UNITER, which is always cross-modal, the margin increases steadily across layers until dropping sharply for the last two. The control task peak for UNITER is similar to that of LXMERT, but the target task peak is lower, suggesting that LXMERT's representations of fine-grained semantic distinctions are slightly more linearly separable than UNITER's. For CLIP, we find that neither target-task nor control-task captions can be linearly separated in the embedding space after the first layer. We investigate the possibility of non-linear representations in \ref{sec:probe4}.

Across all LXMERT layers, target task probes find a linear decision boundary which perfectly separates variants of one caption from variants of the other caption $81.3\%$ of the time, while control task probes able to find a perfect decision boundary only $10.9\%$ of the time.
Among perfect decision boundaries discovered for the target task, the average margin width is $1.9$, while among perfect decision boundaries discovered for the control task, the average margin width is only $0.7$. 

Across UNITER layers, target task probes find a perfect decision boundary 84.7\% of the time with an average margin width of 1.018 while control task probes only do so 19.0\% of the time with an average margin width of 0.48. CLIP target task probes only find a perfect linear decision boundary 3.5\% of the time with an average margin width of only 0.13, while CLIP control task probes never succeed in perfectly separating variant embeddings.

This gap suggests that LXMERT's and UNITER's language layers are \textit{in principle} able to learn easily-extractable representations of Winoground captions, which may capture the differences between semantic minimal pairs. 
Given this result, we ask (in Section~\ref{sec:probe3}) whether using these separable caption sets (as opposed to the original textual minimal pairs) could be used to improve Winoground performance. 
If so, it would suggest that state-of-the-art VL models stumble only on lexically overlapping captions; if not, it would suggest that these models struggle with fine-grained semantic distinctions even in the absence of significant lexical overlap.

\subsection{Do Caption Variants Help with the Winoground Task?}\label{sec:probe3}

To assess whether the separable captions help on the main task, we develop a new test of the Winoground task, using our augmented captions.
Using $n$ different variant-generation methods as defined in Section~\ref{sec:augmentations}, we can obtain sets of captions $T_{0,0}, T_{0,1}, \ldots, T_{0,n-1}$ and $T_{1,0}, T_{1,1}, \ldots, T_{1,n-1}$. 
We then obtain augmentation-aware similarity scores between an image $I$ and a set of $n$ caption variants of $T$, $\{T_1, T_2, \ldots, T_{n}\}$ as $S(I, T) = (1 - \lambda)S(I,T) + \lambda \texttt{agg}(S(I, T_i))$
whereas the similarity score interpolates, with a hyper parameter $\lambda$, between the original similarity score and an aggregate score across all $n$ variants; we experiment with both the max and the mean as possible aggregation functions.
We use these scores to see if performance is improved on Winoground by using caption variants.
We conduct a hyperparameter search over the aggregation function and the value for $\lambda$, as described in Appendix~\ref{app:hyperparameters}.

\begin{table}%
\footnotesize
    \centering
    \setlength{\tabcolsep}{4pt}
    \begin{tabular}{lrrrrrr}
    \toprule
             & \multicolumn{3}{c}{\textbf{Original}} & \multicolumn{3}{c}{\textbf{Augmented}} \\ 
             \cmidrule(lr){2-4}\cmidrule(lr){5-7}
        & \textbf{Text} & \textbf{Img} & \textbf{Grp} & \textbf{Text} & \textbf{Img} & \textbf{Grp} \\
        \cmidrule(lr){2-7}
        Human  & $89.50$ & $88.50$ & $85.50$ & - & - & - \\
LXMERT  & $17.25$ & $5.25$ & $2.75$ & $17.50$ & $4.75$ & $3.25$\\
UNITER & $31.75$ & $10.50$ & $7.25$  &  $31.50$ & $12.50$ & $8.25$ \\
CLIP & $27.50$ & $12.00$ & $9.50$ & $27.25$ & $12.25$ & $9.75$ \\
\bottomrule
    \end{tabular}
    \caption{Results with non-minimal caption variant pairs.}\label{tab:e2}
\end{table}

Results are presented in Table~\ref{tab:e2}. Augmentation does not improve the Text or Group Score by much, indicating that these interventions to increase textual discriminability do not make it easier for the model to pick the correct text given the image. 
This suggests the high lexical similarity between the caption pairs is unlikely to be the main challenge, since models fail to pick between semantically-similar, lexically-different caption candidates. %

\subsection{Distinguishing Captions Conditioned on Caption Variants}\label{sec:probe4}

Finally, we evaluate whether the partitions of the embedding space found in Section \ref{sec:probe2} are meaningful by training MLP probes to select between two captions conditioned on a \textit{different} variant of one of the captions (all paired with the same reference image). These probes, unlike the SVC ones, have the ability to identify and use non-linear patterns in the embeddings.
Intuitively, we are asking: can a probe over the text embeddings produced by each model correctly identify that the caption ``a human viewing a cat on a screen'' is correctly paired with the paraphrase ``a person who looks at a cat on a screen'' (which is a semantic match) and not with a variant of its semantic minimal pair (e.g., ``a cat who looks at a person on a screen'')?
In this experiment we do not train a separate SVC for each Winoground item, but use a single MLP across all Winoground items, increasing the task difficulty significantly. 
If the partitions found for each Winoground item are arbitrary, then a probe trained to distinguish between caption variants should fail on any Winoground item not seen during training.
On the other hand, if a probe is able to distinguish between caption variants for unseen Winoground items, then it must have learned a semantically meaningful partition of the embedding space.
We use the same train/test splits as in Section \ref{sec:probe1}, and a similar control task, in which the labels of a fixed random $50\%$ of Winoground items are swapped.

\paragraph{Results}
Our results are depicted in the second row of Figure~\ref{fig:svc}. 
Performance for LXMERT and UNITER falls between the catastrophic failure of the cross-modal probes in Section \ref{sec:probe1} and the clear success of the unimodal probes in Section \ref{sec:probe2}. 
Performance on the test set is never higher than $60\%$ for any probe size or embedding layer. 
However, target task probes clearly outperform control task probes on test set accuracy, as shown in Figure \ref{fig:svc}. 
On the other hand, MLP probes over CLIP's text branch are more successful than the linear SVC probes over CLIP from Section \ref{sec:probe2}, beating chance by about 10\% accuracy on the test set for the target task. 
This suggests that CLIP may in fact be encoding some semantic distinctions, but that the representations produced by CLIP layers are non-linear.

Test set accuracy clearly improves with layer depth for all three models in early layers, but begins decreasing when cross-modal attention is introduced at layer 9 in LXMERT, and for the final two layers of UNITER. 
This finding mirrors our results from Section \ref{sec:probe2}. The findings for CLIP differ from Section \ref{sec:probe3}, with performance peaking at layer 3 and remaining similar across all subsequent layers.
Performance above chance on this task constitutes some evidence for our hypothesis that text processing is not the primary cause of failure on Winoground for the best current VL models.

\section{Conclusion}

We initially asked whether failures on Winoground occur because SOTA models rely more on bag-of-words than they let on and cannot tell the difference between sentences that contain the same words but differ in meaning.
We found that the story is more complicated: high lexical overlap between captions is not the only---or even the most likely---cause of failure.

First, we showed that it's not only the textual difference between ``a mug in some grass'' and ``a grass in some mug'' that makes Winoground hard. 
Indeed, using Recall@k, we showed that models struggle to identify that \textit{either} minimally different caption matches a particular image.

Next, we re-categorized the Winoground dataset using a set of tags that identify significant challenges beyond semantic compositionally. 
For instance, we identified 38/400 items as \wlab{VisuallyDifficult}, meaning they require identifying a subtle visual feature of the image such as the eye color of a person in an image.
Performance is very low on this subset, for reasons that may have nothing to do with language.
Moreover, some of the images (56/400) and captions (50/400) are unusual or hard to parse: these images and captions are challenging for reasons having nothing to do with their inclusion in a minimal pair.

Even ignoring these cases, we still found that performance on the 171 vanilla Winoground items was low.
To determine whether this is due to the particular zero-shot evaluation setting used by \citet{winoground}, we trained small probes to distinguish between LXMERT or UNITER embeddings of correct matches and incorrect matches. These probes' performance was not consistently better than those trained on a parallel control task, suggesting that zero-shot evaluation is not the source of model failure.

So are these examples hard because models do not understand word order?
We ran a set of experiments in which we made the textual minimal pairs more different from each other: by augmenting the Winoground dataset with variants of each caption, we produced sets of captions which were semantic but not lexical minimal pairs.
Probing the embeddings of these variants, we found that semantic distinctions were linearly separable from LXMERT and UNITER layer representations and non-linearly separable to some extent by LXMERT, UNITER and CLIP representations. Even still, all three models fail to match each set of caption variants with the correct image.
Thus, we observe robust failure on the task even when we use caption variants known to be distinguishable.
It seems that the problem is not simply that the model cannot distinguish between captions with overlapping text, but likely lies in associating those distinctions with images.

Overall, Winoground remains a challenging and promising way to test visuolinguistic ability. 
We would encourage future work to report results on each of the tags we introduce separately, given the clear performance differences across tags we found for CLIP, UNITER, and LXMERT.
And we urge care in drawing conclusions about the compositional abilities of vision-and-language models.

\section*{Limitations}
Like the original Winoground dataset, we evaluate only English. 
Because English is highly word-order dependent, less word-order dependent languages may behave very differently, and, in fact, constructing a Winoground-like dataset in such a language would be non-trivial.
Thus, we should not assume these results generalize to all languages.

We test only 3 types of multimodal models.
While we chose our models to be representative and amenable to the kinds of experiments we were running, we cannot guarantee that our findings apply to all multimodal models.

Also, we focus here mainly on the separability of embeddings in text space. 
There are a parallel set of experiments that could be done for the visual space, but we did not conduct such experiments here. 
Therefore, our conclusions should be limited to what can be concluded from text augmentations.

Finally, we draw some conclusions based on failures to improve models. 
While we believe these negative results are informative, it is of course possible that a better method could be used that would give different results and so one should remain open to this possibility.

\section*{Acknowledgements}

We gratefully thank the Winoground authors for sharing data and helpful conversations, particularly Candace Ross and Adina Williams.
We thank the students in the UT Austin LIN 393 ``What do neural networks know about linguistic structure?'' seminar for input and comments in the early stages of this project.
We also thank Gauri Kambhatla, Vanya Cohen, Jierui Li, and Ray Mooney for helpful discussions.
This work was supported by National Science Foundation Grants No. 2104995 to KM.

\bibliography{anthology,custom}
\bibliographystyle{acl_natbib}

 \appendix

\section{Details on Evaluation Methods}\label{app:eval_traditional_methods}

\subsection{Models}
 Specifically, we use UNITER-base pretrained on COCO Captions, Visual Genome, Conceptual Captions, and SBU Captions as described in \citet{uniter}; LXMERT-base pretrained on COCO Captions, Visual Genome, VQA v2.0, GQA balanced, and VG-QA as descrbed in \citet{lxmert}; and CLIP with a ViT-B/32 image encoder pretrained on WebImageText as described in \citet{clip}.

\section{Method for tagging}\label{app:tags}

The development of these tags occurred in four stages, or ``passes'' through the dataset. In our first pass, we looked briefly at all items to get a broad sense of the dataset, reflecting on and discussing with colleagues any items which caught our interest. The second pass through the dataset looked at each Winoground item carefully one at a time, taking notes on the type of swap performed and any particular challenges or interesting features present in that item. From the $29$ pages of notes produced during the second pass, our final set of $6$ tags was selected to encapsulate broad patterns found throughout the dataset (this number was not fixed in advance, but determined by the number of unique patterns identified). A third pass over the dataset was performed to assign these tags to each image. Finally, since the second and third passes were initially performed by one annotator to ensure consistency, a fourth pass was performed by the other authors to verify the tagging, and Winoground items for which any two annotators disagreed about the tagging were carefully examined and discussed by all annotators to reach a consensus.

After retagging the dataset, the frequency of each new tag was computed, and tag-level performance was measured.

\section{Task Adaptation methods}\label{app:finetune}

We consider the following embeddings as potential inputs to our probes:
\begin{itemize}
    \item Pooled Outputs: The hidden state for the CLS token at the last layer of the model, further processed by a linear projection followed by a Tanh activation.
    \item CLS @ $l$: The hidden state for the CLS token at layer $l$
    \item Mean @ $l$: The vector produced by mean-pooling across all hidden states at layer $l$
    \item Max @ $l$: The vector produced by max-pooling across all hidden states at layer $l$
\end{itemize}

Let the function $E(I,T)$ be application of a given model to input image $I$ and caption $T$, followed by extraction of the embedding being probed over. Our probes are then given a pair of concatenated embeddings for two candidate image-caption pairs, and asked to output $0$ if the first pair is a better match or $1$ if the second pair is a better match. Specifically, we define the following task corresponding to Text Score for a probe $P(\cdot)$:
\begin{align*}
    P\big(E(I_0,T_0)\Vert E(I_0,T_1)\big) &\rightarrow 0\\
    P\big(E(I_0,T_1)\Vert E(I_0,T_0)\big) &\rightarrow 1\\
    P\big(E(I_1,T_0)\Vert E(I_1,T_1)\big) &\rightarrow 1\\
    P\big(E(I_1,T_1)\Vert E(I_1,T_0)\big) &\rightarrow 0
\end{align*}
An equivalent probing task corresponding to Image Score is defined:
\begin{align*}
    P\big(E(I_0,T_0)\Vert E(I_1,T_0)\big) &\rightarrow 0\\
    P\big(E(I_1,T_0)\Vert E(I_0,T_0)\big) &\rightarrow 1\\
    P\big(E(I_0,T_1)\Vert E(I_1,T_1)\big) &\rightarrow 1\\
    P\big(E(I_1,T_1)\Vert E(I_0,T_1)\big) &\rightarrow 0
\end{align*}

Our control task is formulated nearly identically to the target task, except that a random 50\% of the Winoground items in each split are chosen at the start of training, and the labels for these items are flipped. That is, if our target task has the label $0$ and the given item is not flipped, the control task also has the label $0$; if our target task has the label $0$ and the given item is flipped, the control task has the label $1$. 

Since the flipped items are selected before training, a probe which simply memorizes the data will perform well on this control task for items it has already seen. We therefore split the data into a training and a testing set, the latter of which is never seen during training. This means we are dividing the $400$ Winoground items into four new splits: $150$ training samples whose labels are not flipped in the control task, $150$ training samples whose labels are flipped, $50$ testing samples whose labels are not flipped, and $50$ testing samples whose labels are. In order to ensure each split is representative of overall the Winoground benchmark, we perform stratified sampling, where are buckets are any combination of the ``Pragmatics'', ``Symbolic'', and ``Morpheme-Level'' visual tags and the ``Both'' linguistic tag. We subsequently confirm that the ratio of each visual and linguistic tag, as well as of each new tag introduced here, is similar across each split.

We test a variety of small Multi-Layer Perceptron (MLP) probes over these extracted embeddings, each of which maps from the input dimension of $2\times768$ to a single output. ReLU activation is applied at intermediate layers, and Sigmoid activation is used at the final layer to ensure the output is in the range $[0,1]$. We empirically select a hidden size of $1024$, a learning rate of $0.0001$, $4$ MLP layers, and $200$ epochs of training to use for every probe, after ablating each of these hyperparameters for all combinations of probe and embedding types.

We use a single NVIDIA RTX 8000 GPU for all our experiments. All probes took no more than $10$ hours to run.

\section{Textual Variants Methods}
\label{app:variants}

\subsubsection{Syntactic augmentations}
We generate a maximum of $3$ variants using the \texttt{PropbankSRLRoles} augmentation. 
This augmentation extracts semantic role labels for the provided sentence using the AllenNLP implementation~\footnote{\url{https://demo.allennlp.org/semantic-role-labeling}} of SRL BERT~\citep{https://doi.org/10.48550/arxiv.1904.05255} and applies its hardcoded syntactic rules (if applicable) to generate a new sentence.

\subsubsection{Semantic, word-based augmentations}
We use $4$ different augmentation methods, each of which randomly replace words in the sentence with new approximately meaning-preserving words. All methods use SpaCy~\citep{Honnibal_spaCy_Industrial-strength_Natural_2020} to parse the sentence to perform POS tagging.

\noindent \texttt{ReplaceHyponyms}, \texttt{ReplaceHypernyms}. The first augmentation replaces a noun with a hyponym and the second replaces a noun with a hypernym. We generate a maximum of $2$ variants per augmentation. This method uses CheckList~\citep{ribeiro-etal-2020-beyond} for the list of hyponyms/hypernyms.

\noindent \texttt{Slangificator}. It replaces a word with a slang word. This uses a manually curated list of word -> slang word mappings. We generate a maximum of $3$ variants.

\noindent \texttt{SynonymSubstitution}. It replaces a word with a synonym based on WordNet~\citep{miller1998wordnet} via NLTK~\citep{bird2006nltk}. We generate a maximum of $3$ variants.

\subsubsection{Paraphrasing}
\noindent \texttt{Backtranslation}. It translates a sentence to German and back using FSMT~\citep{ng-etal-2019-facebook}. We generate a maximum of $1$ variant.

\noindent \texttt{DiverseParaphrase}. It generates diverse paraphrases using DiPS~\citep{kumar-etal-2019-submodular} equipped with Diverse Beam Search~\citep{AAAI1817329}. We generate a maximum of $3$ variants.

\noindent \texttt{ProtAugmentDiverseParaphrase}. It generates diverse paraphrases using ProtAugment~\citep{dopierre-etal-2021-protaugment}. We generate a maximum of $5$ variants.

\subsubsection{Identity}
We also define the original input text as a `variant' that has undergone the identity transformation.

\subsection{Discriminable Caption Pair Experiment}\label{app:hyperparameters}

To assess whether the separable captions help on the main task, we develop a new test of the Winoground task, using our augmented captions.
Using $n$ different variant-generation methods as defined in Section~\ref{sec:augmentations}, we can obtain sets of captions $T_{0,0}, T_{0,1}, \ldots, T_{0,n-1}$ and $T_{1,0}, T_{1,1}, \ldots, T_{1,n-1}$. 
Then, every multimodal model under consideration outputs a similarity score  $\simm(I, T)$ given an image $I$ and text $T$ as input. We define augmentation-aware similarity scores between a given image $I$ and a set of $n$ caption variants of $T$, $\{T_0, T_1, \ldots, T_{n-1}\}$ as follows:
\begin{align*}
    \simm_{aug, mean}(I,T) = &(1- \lambda) \simm(I,T_0) + \\ &\lambda \meann_i [\simm(I, T_i)] \\
    \simm_{aug, max}(I,T) = &(1- \lambda) \simm(I,T_0) + \\ &\lambda \max_i [\simm(I, T_i)]
\end{align*}
\vspace{-.1in}

where the choice of using $\simm_{aug, mean}$ vs. $\simm_{aug, max}$ and the value of $\lambda$ are hyperparameters. This similarity score interpolates, using $\lambda$, between the original similarity score and an aggregated (max/mean) score across all $n$ variants. We can use these scores to see if performance is improved on Winoground by using caption variants.

We conduct a hyperparameter search over the similarity function and the value for $\lambda$. We found that $\simm_{aug, mean}$ works best for LXMERT while $\simm_{aug, max}$ works best for UNITER and CLIP. We picked the best value for $\lambda$ by testing every $\lambda$ value between $0$ and $1$ in steps of $0.25$ and picking the value that maximizes the group score. $\lambda=0.5$ works best for LXMERT, $\lambda=0.75$ for UNITER and CLIP.

\section{Winoground: New Tags}
\label{sec:appendixnewtags}
Our new tags appear in Table~\ref{tab:tags}. For the full Winoground dataset, see \url{https://huggingface.co/datasets/facebook/winoground}.

\begin{table*}[ht]
    \footnotesize
    \centering
    \begin{tabular}{p{1.3in}|p{4.4in}}
        \textbf{Tag} & \textbf{Winoground Items} \\ \hline
         \wlab{NonCompositional} & 72, 73, 74, 95, 96, 133, 149, 150, 164, 218, 221, 222, 224, 235, 237, 246, 274, 275, 321, 325, 326, 327, 332, 333, 334, 350, 364, 365, 398, 399 \\
         \wlab{AmbiguouslyCorrect} & 3, 13, 36, 46, 75, 76, 77, 78, 82, 86, 88, 113, 119, 121, 131, 132, 133, 148, 189, 220, 221, 262, 263, 287, 295, 300, 303, 305, 307, 310, 319, 322, 332, 340, 343, 344, 348, 353, 355, 356, 363, 374, 377, 381, 386, 394 \\
         \wlab{VisuallyDifficult} & 4, 22, 23, 25, 27, 28, 31, 36, 58, 65, 69, 70, 77, 97, 116, 118, 134, 138, 159, 163, 172, 176, 182, 187, 200, 214, 226, 227, 232, 241, 255, 268, 286, 335, 352, 356, 373, 376 \\
         \wlab{UnusualImage} & 31, 36, 38, 41, 42, 61, 62, 70, 78, 84, 93, 110, 114, 116, 128, 133, 136, 155, 159, 164, 173, 174, 188, 201, 203, 204, 206, 209, 218, 223, 239, 245, 246, 247, 254, 274, 275, 277, 280, 282, 293, 303, 307, 314, 319, 320, 327, 329, 339, 362, 367, 383, 384, 388, 393, 395 \\
         \wlab{UnusualText} & 10, 41, 49, 58, 63, 68, 70, 152, 156, 159, 163, 174, 198, 201, 209, 214, 215, 221, 229, 233, 237, 253, 257, 264, 275, 287, 303, 315, 318, 323, 324, 326, 327, 335, 338, 342, 343, 345, 346, 351, 354, 359, 364, 376, 383, 385, 386, 387, 390, 394 \\
         \wlab{ComplexReasoning} & 16, 40, 44, 46, 55, 58, 81, 83, 93, 97, 103, 111, 116, 118, 128, 130, 135, 143, 144, 176, 190, 191, 192, 193, 199, 206, 208, 209, 210, 211, 217, 218, 219, 227, 228, 230, 234, 238, 241, 242, 249, 254, 258, 260, 262, 264, 267, 268, 275, 276, 281, 284, 286, 287, 292, 295, 296, 298, 299, 304, 311, 312, 316, 330, 331, 334, 336, 342, 347, 358, 361, 371, 373, 375, 382, 383, 392, 396\\
         \wlab{NoTag} & 0, 1, 2, 5, 6, 7, 8, 9, 11, 12, 14, 15, 17, 18, 19, 20, 21, 24, 26, 29, 30, 32, 33, 34, 35, 37, 39, 43, 45, 47, 48, 50, 51, 52, 53, 54, 56, 57, 59, 60, 64, 66, 67, 71, 79, 80, 85, 87, 89, 90, 91, 92, 94, 98, 99, 100, 101, 102, 104, 105, 106, 107, 108, 109, 112, 115, 117, 120, 122, 123, 124, 125, 126, 127, 129, 137, 139, 140, 141, 142, 145, 146, 147, 151, 153, 154, 157, 158, 160, 161, 162, 165, 166, 167, 168, 169, 170, 171, 175, 177, 178, 179, 180, 181, 183, 184, 185, 186, 194, 195, 196, 197, 202, 205, 207, 212, 213, 216, 225, 231, 236, 240, 243, 244, 248, 250, 251, 252, 256, 259, 261, 265, 266, 269, 270, 271, 272, 273, 278, 279, 283, 285, 288, 289, 290, 291, 294, 297, 301, 302, 306, 308, 309, 317, 328, 337, 341, 349, 357, 360, 366, 368, 369, 370, 372, 378, 379, 380, 389, 391, 397 

    \end{tabular}
    \caption{Winoground Tags with which items fall into each number, using numbering scheme from \cite{winoground}}
    \label{tab:tags}
\end{table*}

\end{document}